\documentclass[sigconf, screen]{acmart}

\usepackage{amsmath}
\usepackage{subcaption}
\usepackage{multirow}
\usepackage{pifont}
\usepackage{makecell}
\usepackage{enumitem}
\usepackage{algorithm}
\usepackage{algorithmic}

\AtBeginDocument{%
  }

\copyrightyear{2025}
\acmYear{2025}
\setcopyright{acmlicensed}
\acmConference[MM '25]{Proceedings of the 33rd ACM International Conference on Multimedia}{October 27--31, 2025}{Dublin, Ireland}
\acmBooktitle{Proceedings of the 33rd ACM International Conference on Multimedia (MM '25), October 27--31, 2025, Dublin, Ireland}
\acmDOI{XX.XXXX/XXXXXXX.XXXXXXX}
\acmISBN{XXX-X-XXXX-XXXX-X/XX/XX}

\begin{document}

\title{Fast3D: Accelerating 3D Multi-modal Large Language Models for Efficient 3D Scene Understanding}

\author{Wencan Huang}
\email{huangwencan@stu.pku.edu.cn}
\affiliation{%
  \institution{Wangxuan Institute of Computer Technology, Peking University}
  \city{Beijing}
  \country{China}
}

\author{Daizong Liu}
\email{dzliu@stu.pku.edu.cn}
\affiliation{%
  \institution{Wangxuan Institute of Computer Technology, Peking University}
  \city{Beijing}
  \country{China}
}

\author{Wei Hu}
\authornote{This paper is supported by the State Key Laboratory of General Artificial Intelligence. Corresponding author: Wei Hu (forhuwei@pku.edu.cn).}
\email{forhuwei@pku.edu.cn}
\affiliation{%
  \institution{Wangxuan Institute of Computer Technology, Peking University}
  \city{Beijing}
  \country{China}
}

\renewcommand{\shortauthors}{Wencan Huang, Daizong Liu, \& Wei Hu}

\begin{abstract}
While 3D Multi-modal Large Language Models (MLLMs) demonstrate remarkable scene understanding capabilities, their practical deployment faces critical challenges due to computational inefficiency. The key bottleneck stems from processing excessive object-centric visual tokens required for comprehensive 3D scene representation. Although visual token pruning has shown promise in accelerating 2D MLLMs, its applicability to 3D domains remains largely unexplored due to fundamental disparities in token structures. In this paper, we reveal two critical insights: (1) Significant redundancy exists in object-level 3D token representations, analogous to patch-level redundancy in 2D systems; (2) Global attention patterns exhibit strong predictive power for identifying non-essential tokens in 3D contexts. Building on these observations, we propose Fast3D, a plug-and-play visual token pruning framework for 3D MLLMs featuring two technical innovations: (1) Global Attention Prediction (GAP), where a lightweight neural network learns to predict the global attention distributions of the target model, enabling efficient token importance estimation for precise pruning guidance; (2) Sample-Adaptive visual token Pruning (SAP), which introduces dynamic token budgets through attention-based complexity assessment, automatically adjusting layer-wise pruning ratios based on input characteristics. Both of these two techniques operate without modifying the parameters of the target model. Extensive evaluations across five benchmarks validate the effectiveness of Fast3D, particularly under high visual token pruning ratios. Code is available at \url{https://github.com/wencan25/Fast3D}.
\end{abstract}

\begin{CCSXML}
<ccs2012>
    <concept>
        <concept_id>10010147.10010178</concept_id>
        <concept_desc>Computing methodologies~Artificial intelligence</concept_desc>
        <concept_significance>500</concept_significance>
    </concept>
    <concept>
        <concept_id>10010147.10010178.10010224</concept_id>
        <concept_desc>Computing methodologies~Computer vision</concept_desc>
        <concept_significance>500</concept_significance>
    </concept>
    <concept>
        <concept_id>10010147.10010178.10010179</concept_id>
        <concept_desc>Computing methodologies~Natural language processing</concept_desc>
        <concept_significance>500</concept_significance>
    </concept>
</ccs2012>
\end{CCSXML}

\ccsdesc[500]{Computing methodologies~Artificial intelligence}
\ccsdesc[500]{Computing methodologies~Computer vision}
\ccsdesc[500]{Computing methodologies~Natural language processing}

\keywords{Multi-modal Large Language Models, Visual Token Pruning, 3D Scene Understanding}


\maketitle

\begin{figure}[!t]
\centering
\includegraphics[width=0.97\linewidth]{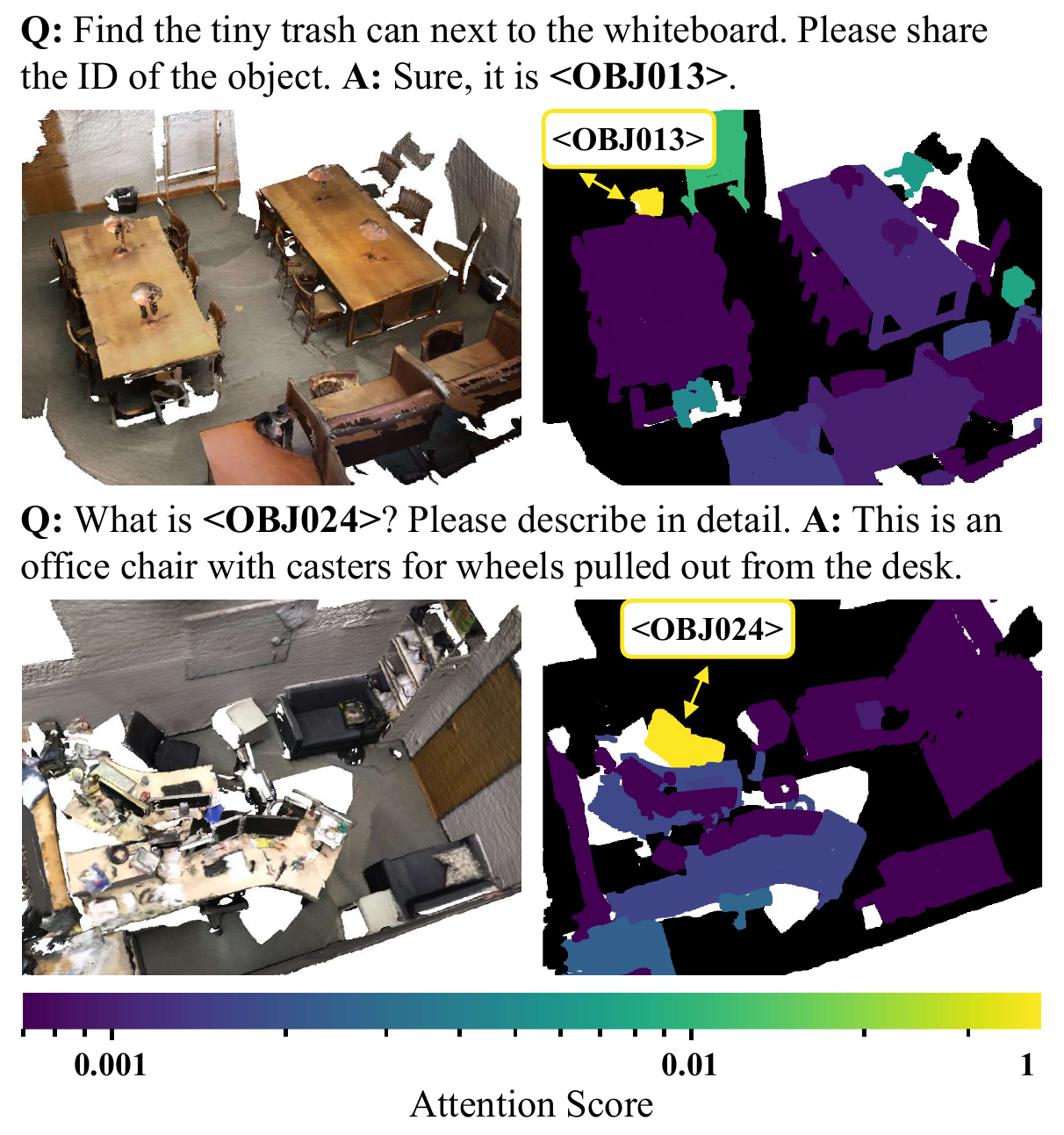}
\vspace{-2mm}
\caption{Visualization of the global attention maps aggregated from all layers of Chat-Scene \cite{huang2024chat}, highlighting significant concentration of attention weights on text-relevant 3D objects, which indicates the viability of attention-guided visual token pruning for object-centric 3D MLLMs.}
\vspace{-4mm}
\label{fig:intro}
\end{figure}
\section{Introduction}

The remarkable success of Large Language Models (LLMs) \cite{chiang2023vicuna,achiam2023gpt,touvron2023llama} and Multi-modal LLMs (MLLMs) \cite{liu2024visual,zhu2023minigpt,lai2024lisa} has sparked growing interest in extending their applications to the 3D domain for embodied intelligent agents. By encoding object-centric 3D representations and mapping these features onto the latent semantic space of LLMs as visual tokens, 3D MLLMs \cite{huang2023embodied,huang2024chat,kang2024robin3d,zemskova20243dgraphllm} have demonstrated exceptional scene understanding capabilities in core 3D vision-language tasks such as referred object grounding \cite{chen2020scanrefer,zhang2023multi3drefer}, dense scene captioning \cite{chen2021scan2cap}, and situated question answering \cite{azuma2022scanqa,ma2022sqa3d}, leveraging a unified multi-modal sequence modeling paradigm.

However, this paradigm often suffers from substantial computational overhead due to the reliance on an excessive number of visual tokens, limiting practical deployment. For instance, Chat-Scene \cite{huang2024chat} adopts 300 visual tokens for a 100-object scene through tripartite encoding, which requires over 5 times the token consumption compared to its text-only inference on ScanQA \cite{azuma2022scanqa}. More recent approaches \cite{zemskova20243dgraphllm} resort to integrating auxiliary visual tokens such as semantic relationships between objects to enhance performance, inflating the visual token count to 800 per scene, which further exacerbates the already high computation of 3D MLLMs.

Training-free visual token pruning approaches \cite{bolya2022token,shang2024llava,chen2025image,zhang2024sparsevlm,zhao2024stitch} have emerged as a compelling solution for accelerating 2D MLLMs. Previous methods \cite{chen2025image,zhang2024sparsevlm} typically utilize partial model information (\textit{e.g.}, attention maps from early layers) to eliminate less critical image tokens. Recent advances like SGL \cite{zhao2024stitch} demonstrate enhanced efficiency by leveraging global attention patterns from compact MLLMs to guide token pruning in larger counterparts. While these techniques have significantly optimized 2D MLLMs, their applicability to 3D MLLMs remains largely unexplored due to fundamental differences in token structures. Unlike the inherently redundant patch-level tokens in 2D vision systems, the visual tokens in representative 3D MLLMs are object-centric with explicit spatial-semantic grounding. This structural distinction raises critical questions: \textit{Does conventional token redundancy persist in 3D representations? Can attention-guided pruning maintain effectiveness when applied to discrete object tokens?} To investigate these questions, we conduct a preliminary study by visualizing the global attention distributions in Chat-Scene \cite{huang2024chat}. As shown in Figure \ref{fig:intro}, we observe that the aggregated attention maps accurately concentrate on a few text-relevant objects, demonstrating significant redundancy even in structured 3D representations. Our empirical evidence reveals two key findings: \textit{(1) Object-centric architectures do not inherently eliminate token redundancy, and (2) Global attention patterns aggregated from all layers of the target model are superior pruning indicators for 3D MLLMs.}

However, retrieving attention maps from all layers necessitates full-model inference, making it infeasible for practical deployment. To mitigate this, existing approaches like SGL \cite{zhao2024stitch} attempt to leverage the aggregated attention maps from a smaller MLLM to guide pruning in a larger one. But this method suffers from two fundamental limitations. First, the smaller model is trained independently without joint optimization, leading to potential inconsistencies in attention distributions. Second, the smaller model in SGL, with its 2 billion parameters, is not sufficiently lightweight, introducing non-negligible computational overhead. These issues are further compounded in resource-constrained 3D domains, where such small-scale yet capable MLLMs may not be readily available. Therefore, how to accurately identify critical visual tokens in 3D MLLMs with minimal computation remains an open challenge.

To address this challenge, we present \textbf{Fast3D}, a plug-and-play visual token pruning framework to reduce the inference cost of 3D MLLMs with two technical innovations. First, we develop a \textit{Global Attention Prediction (GAP)} mechanism that systematically identifies essential 3D visual tokens through attention distillation. Specifically, we propose training a lightweight GAP network to predict the oracle attention map aggregated across all layers of 3D MLLMs capturing three crucial interactions: the self-attention of vision tokens and their cross-modal interactions with prompt and generated text. The network adopts an encoder-decoder architecture with early fusion of three complementary embeddings -- visual semantics, spatial coordinates, and object identifiers -- resulting in reduced computational cost and improved prediction accuracy. During inference, the predicted attention maps provide a precise guidance for pruning less important visual tokens in 3D MLLMs.

While efficient token importance estimation is essential, when and how to prune visual tokens remains a key challenge. Existing pruning strategies \cite{chen2025image,zhao2024stitch,ye2024fit} typically rely on static, empirically set pruning ratios, applying uniform token budgets across all inputs regardless of task complexity. Such rigidity ignores the inherent diversity in visual-linguistic difficulty across samples. To overcome this limitation, we introduce \textit{Sample-Adaptive visual token Pruning (SAP)}, which dynamically adjusts token budgets based on attention-driven complexity assessment. SAP estimates the per-sample difficulty and adaptively allocates pruning ratios at each transformer layer, allowing the model to retain more tokens for complex instructions and fewer for simpler ones. To further enhance deployment efficiency, we perform an offline binary search to optimize the pruning configuration under fixed computational budgets, achieving improved overall accuracy-efficiency trade-offs.

In summary, our main contributions are threefold:

\begin{itemize}
[leftmargin=0.4cm, itemindent=0cm]
    \item We reveal the substantial redundancy of visual tokens in object-centric 3D MLLMs via investigating their global attention patterns, indicating that a great number of visual tokens can be discarded during inference with negligible performance drop.
        
    \item We propose \textbf{Fast3D}, a plug-and-play visual token pruning framework tailored for 3D MLLMs featuring two novel components: (1) \emph{Global Attention Prediction (GAP)}, where we train a lightweight neural network for efficient token importance estimation; (2) \emph{Sample-Adaptive visual token Pruning (SAP)}, which dynamically adjusts token budgets based on input complexity to achieve improved overall accuracy-efficiency trade-offs.

    \item We demonstrate the superiority of Fast3D through comprehensive evaluations across five 3D vision-language tasks, enabling up to 90\% visual token reduction in 3D MLLMs like Chat-Scene while maintaining over 96.8\% of original performance, achieving significantly higher efficiency for 3D scene understanding.
\end{itemize}
\section{Related Work}

\subsection{3D Multi-modal Large Language Models}

The prevalence of Large Language Models (LLMs) \cite{chiang2023vicuna,touvron2023llama} and Multi-modal LLMs (MLLMs) \cite{liu2024visual,zhu2023minigpt,wang2024gpt4video,liu2024surveyA,liu2024pandora} has sparked growing interest in extending their reasoning capabilities to the 3D domain, leading to the emergence of 3D MLLMs for point cloud object understanding \cite{tang2024minigpt,qi2024shapellm,xu2024pointllm,guo2023point,han2023imagebind,qi2024gpt4point,chu2025daffordancellm} and 3D scene understanding \cite{huang2024chat,hong20233d,he2024segpoint,chen2024ll3da,fu2024scene,chen2024grounded,qi2025gpt4scene,liu2024survey,huang2023dense,huang2024advancing,yu2025inst3d,wang2025data,liu2025seeing}. For 3D scene-level MLLMs, the pioneering work \cite{hong20233d} renders 3D environments into multi-view images, leveraging pre-trained 2D MLLMs for 3D comprehension. Subsequent works directly project scene-level point clouds into the embedding space of LLMs via MLPs \cite{he2024segpoint} or Qformers \cite{chen2024ll3da} for enhanced structural preservation. Recent models such as Chat-3D \cite{wang2023chat,huang2023chat}, Chat-Scene \cite{huang2024chat}, LEO \cite{huang2023embodied}, Robin3D \cite{kang2024robin3d}, and 3DGraphLLM \cite{zemskova20243dgraphllm} predominantly adopt object-centric architectures to capture intricate spatial-semantic relationships. For example, Chat-3D v2 \cite{huang2023chat} employs off-the-shelf 3D detectors \cite{schult2023mask3d} to extract per-object spatial features and introduces object identifiers for precise object referencing and grounding. More recent approaches further enhance performance by integrating auxiliary visual tokens, such as multi-view 2D representations \cite{huang2024chat} and semantic relationships between objects \cite{zemskova20243dgraphllm}. However, the increasing number of visual tokens also greatly exacerbates the already high computational costs of 3D MLLMs. In this paper, we address the token overhead with a plug-and-play visual token pruning framework, enabling more efficient 3D scene understanding.

\subsection{Visual Token Compression}

Visual token compression has been widely explored in transformer-based vision models, such as ViTs \cite{vaswani2017attention,alexey2020image}, aimed at reducing computational overhead through token pruning \cite{liang2022not,rao2021dynamicvit,fayyaz2022adaptive,wang2024zero,cao2024madtp}, merging \cite{bolya2022token,chen2023diffrate,shi2023crossget,ju2024turbo}, and skipping \cite{han2021dynamic,han2024latency,meng2022adavit,zhao2024dynamic,zhao2024dynamictuning} techniques. In the context of MLLMs, prior works have explored compression strategies from both efficient projector design \cite{li2024llama,cha2024honeybee,li2024tokenpacker,chu2023mobilevlm,chu2024mobilevlm} and dynamic token reduction \cite{ye2024voco,yao2024deco,meng2024deepstack,chen2024llavolta} perspectives, seeking to eliminate redundant visual tokens while maintaining model performance. However, these methods often require additional fine-tuning of the base model. In contrast, more recent training-free approaches \cite{shang2024llava,chen2025image,zhang2024sparsevlm,huang2024ivtp,zhao2024stitch,zhong2024aim,wang2024cls,zhuang2024st,jiang2025kind,zhao2024accelerating,han2024rethinking,ye2024fit,lin2024boosting,li2025beyond} can be deployed directly on pre-trained MLLMs without requiring parameter updates, offering a more practical solution for improving efficiency. For instance, methods like LLaVA-PruMerge \cite{shang2024llava} and VTC-CLS \cite{wang2024cls} merge tokens within the visual encoder before feeding them into the LLM. Other approaches, such as FastV \cite{chen2025image} and Sparsevlm \cite{zhang2024sparsevlm}, prune inattentive visual tokens in specific LLM layers based on attention maps between visual and instruction tokens, utilizing manually designed reduction strategies. Meanwhile, techniques like FitPrune \cite{ye2024fit} attempt to automatically determine the optimal pruning recipe under predefined computation budgets. However, these methods can struggle to accurately identify essential visual tokens due to their reliance on partial information from early LLM layers. Recently, SGL \cite{zhao2024stitch} shows that the global attention map aggregated from all layers of small MLLMs can provide better guidance for pruning in larger MLLMs. Despite these advancements, limited research explores the effectiveness of visual token pruning for object-centric architectures like 3D MLLMs.

\section{Methodology}

We begin by providing the preliminaries on typical scene-level 3D MLLMs in Section \ref{sec:pre}. Subsequently, we detail Fast3D, the proposed plug-and-play visual token pruning framework, which incorporates two innovative techniques. Specifically, Section \ref{sec:gap} details the Global Attention Prediction (GAP) mechanism for efficient token importance estimation, while Section \ref{sec:sap} introduces the Sample-Adaptive visual token Pruning (SAP) strategy.

\subsection{Preliminary}
\label{sec:pre}

In this study, we use Chat-Scene \cite{huang2024chat}, a state-of-the-art 3D MLLM with an exemplar object-centric architecture for 3D scene understanding, as the base model for demonstrating our proposed inference acceleration framework. Chat-Scene features three core components: 1) Visual Encoding: This module processes raw 3D scenes into structured object-level representations through tripartite encoding. First, a pre-trained Mask3D detector \cite{schult2023mask3d} decomposes the scene into $n$ object proposals. Then, each object receives a unique object identifier embedding $\mathbf{E}^o_i \in \mathbb{R}^d$ added to the language model vocabulary, the geometric embedding $\mathbf{E}^p_i \in \mathbb{R}^{d_p}$ from point clouds via Uni3D \cite{zhou2023uni3d}, and the visual embedding $\mathbf{E}^v_i \in \mathbb{R}^{d_v}$ extracted from multi-view images using DINOv2 \cite{oquab2023dinov2}. 2) Visual Projection: Object embeddings are mapped to the LLM's latent space as visual tokens. For the $i$-th object, its visual tokens are represented as $\mathbf{V}^o_i = \mathbf{E}^o_i$, $\mathbf{V}^p_i = f_p(\mathbf{E}^p_i)$, and $\mathbf{V}^v_i = f_v(\mathbf{E}^v_i)$, where $f_p(\cdot)$ and $f_v(\cdot)$ are fully connected projectors. 3) Multi-modal Language Modeling: The LLM takes visual and textual tokens as input, and generates the next text token in an autoregressive manner. At inference step $i$, the input sequence is:
\begin{gather}
\mathbf{X}^i = [\mathbf{V}, \mathbf{T}, \mathbf{G}^i] \in \mathbb{R}^{(3n+m+i)\times d}, \\
\mathbf{V} = \Vert_{k=1}^n [\mathbf{V}^o_k, \mathbf{V}^p_k, \mathbf{V}^v_k] \in \mathbb{R}^{3n\times d},
\end{gather}
where $\mathbf{V} \in \mathbb{R}^{3n\times d}$ denotes the concatenated visual tokens, $\mathbf{T} \in \mathbb{R}^{m\times d}$ is the textual prompt tokens, and $\mathbf{G}^i$ is the previous generated text tokens. With tripartite encoding, we usually have $3n \gg m$, especially for complex 3D scenes with dense objects.

\begin{figure*}[!t]
    \centering
    \includegraphics[width=1\linewidth]{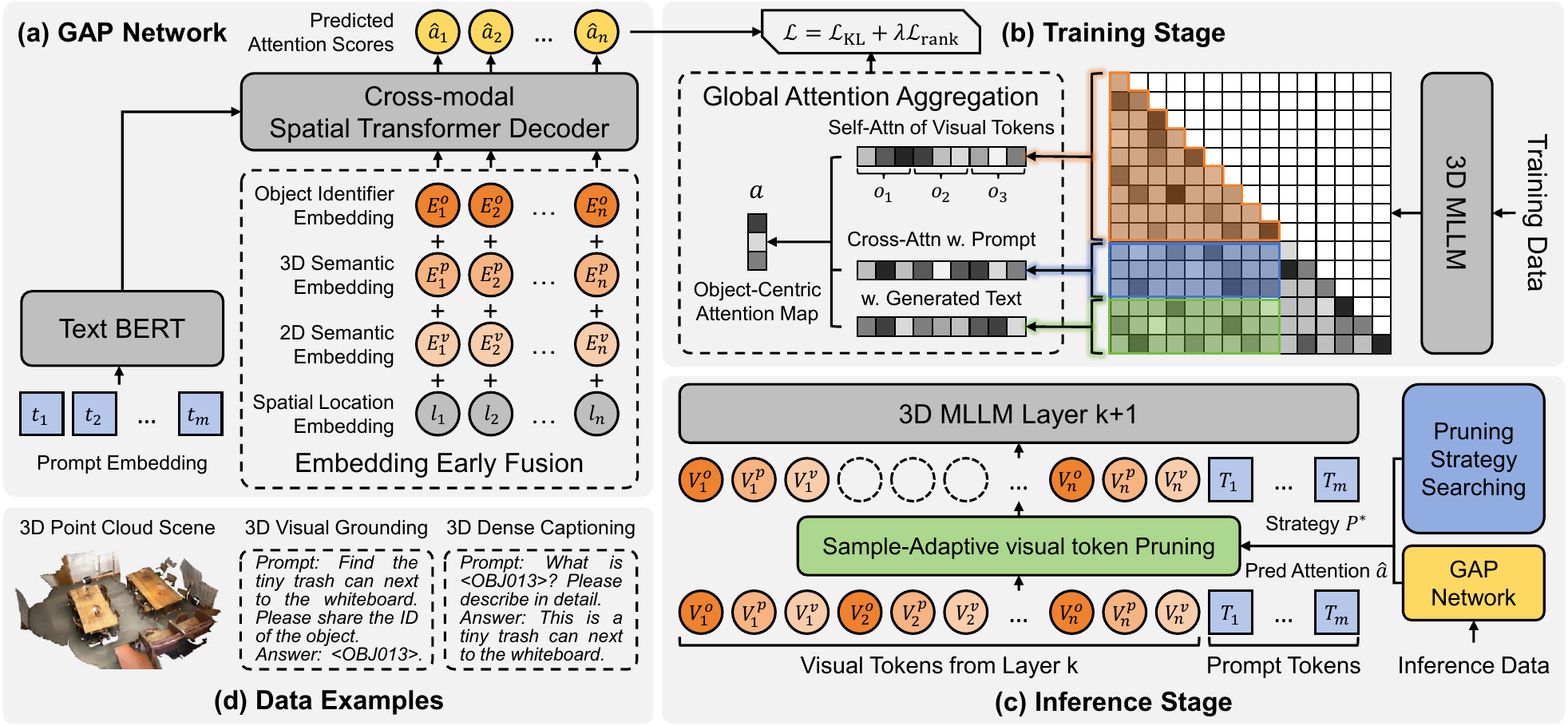}
    \vspace{-4mm}
    \caption{Framework of Fast3D. (a) The GAP network learns to predict the aggregated attention maps of the target 3D MLLM, providing efficient and accurate guidance for visual token pruning. (b) During training, we construct the ground-truth object-centric attention map by aggregating attention scores of visual tokens received from other tokens across all heads and layers of the target model. (c) At inference stage, we employ an optimized pruning strategy to eliminate less-important visual tokens sample-adaptively based on GAP predictions. (d) Data examples of diverse 3D scene understanding tasks.}
    \label{fig:method}
\end{figure*}

\subsection{Global Attention Prediction}
\label{sec:gap}

The computational burden of processing excessive visual tokens significantly impedes 3D MLLM inference efficiency. While attention-guided visual token pruning \cite{chen2025image,zhao2024stitch} shows promise in accelerating 2D MLLMs, its adaptation to object-centric 3D MLLMs remains unexplored. Our empirical analysis demonstrates that \emph{global attention patterns aggregated from all layers of the target model retain the viability as superior pruning indicators in 3D domains} (Table \ref{table:attention_map_source}). However, deriving these oracle attention maps requires full-model inference, creating a practical deployment paradox. Existing approximations like SGL \cite{zhao2024stitch} that employs smaller MLLMs to guide pruning in larger models exhibit fundamental limitations: 1) Attention pattern mismatch due to independent model training without joint optimization, 2) Non-negligible computational overhead from insufficiently compact proxy models, and 3) Scarcity of pre-trained small MLLMs in 3D domains. This presents a critical challenge:

\emph{How to accurately identify essential visual tokens in 3D MLLMs with minimal computation?}

We introduce Global Attention Prediction (GAP) (Figure \ref{fig:method}a-b) as a novel solution: \emph{training a lightweight GAP network to predict the aggregated attention map} of the target 3D MLLM as a precise and efficient pruning guidance.

\subsubsection{\textbf{Global Attention Aggregation.}}
In order to train the GAP network, we extract ground-truth global attention maps from the frozen target 3D MLLM, as shown in Figure \ref{fig:method}b. Given an input training sequence $\mathbf{X}$ composed of $3n$ visual tokens, $m$ prompt tokens, and $t$ generated tokens, let $\mathbf{A}^{(l,h)}\in \mathbb{R}^{(3n+m+t) \times (3n+m+t)}$ denote the attention matrix at layer $l$ and head $h$. The aggregated attention map is computed via uniform averaging across all $L$ layers and $H$ heads:
$\mathbf{A} = \frac{1}{LH} \sum_{l=1}^L \sum_{h=1}^H \mathbf{A}^{(l,h)} \in \mathbb{R}^{(3n+m+t)\times (3n+m+t)}.$
We then aggregate the attention scores of visual tokens received from other tokens to form a global object-centric attention map through three complementary mechanisms:

\noindent{\textbf{Self-Attention of Visual Tokens.}} Let $\mathbf{A}_{\text{self}} \in \mathbb{R}^{3n\times 3n}$ denote the top-left submatrix of $\mathbf{A}$, where each entry $A_{\text{self}}(i,j)$ reflects the dependency between visual tokens $\mathbf{V}_i$ and $\mathbf{V}_j$. Considering the causal attention structure, we compute visual token importance $\boldsymbol{a}_{self} \in \mathbb{R}^{3n}$ using masked column-wise averaging:
\begin{gather}
a_{\text{self}}(j) = \frac{1}{3n-j+1}\sum_{i=1}^{3n} M_{\text{causal}}(i,j) \cdot A_{\text{self}}(i,j),
\end{gather}
where $\mathbf{M}_{\text{causal}}$ is a lower-triangular mask ensuring visual token $\mathbf{V}_i$ only attends to preceding tokens $\mathbf{V}_{j \leq i}$. High-value entries in $\boldsymbol{a}_{\text{self}}$ indicate geometrically or semantically correlated objects that the 3D MLLM inherently prioritizes for visual reasoning.

\noindent{\textbf{Cross-Attention with Prompt.}} For $m$ prompt tokens, let $\mathbf{A}_{\text{prompt}} \in \mathbb{R}^{m \times 3n}$ be the cross-attention sub-matrix. We derive prompt-aware token importance $\boldsymbol{a}_{\text{prompt}} \in \mathbb{R}^{3n}$ through average pooling:
\begin{gather}
    a_{\text{prompt}}(j) = \frac{1}{m}\sum_{i=1}^{m} A_{\text{prompt}}(i,j),
\end{gather}
which highlights visual tokens most responsive to task-specific linguistic cues, ensuring instruction-aligned feature preservation.

\noindent{\textbf{Cross-Attention with Generated Text.}} For $t$ generated text tokens with teacher forcing, let $\mathbf{A}_{\text{text}} \in \mathbb{R}^{t \times 3n}$ denote the bottom-left block. We compute generation-aware importance $\boldsymbol{a}_{\text{text}} \in \mathbb{R}^{3n}$ with a confidence-weighted average:
\begin{equation}
{a}_{\text{text}}(j) = \frac{1}{\sum_{i=1}^t {s}^{(i)}} \sum_{i=1}^{t} {s}^{(i)} \cdot A_{\text{text}}(i,j),
\end{equation}
where $s^{(i)}$ represents the softmax probability of the next generated token at the $i$-th step. This emphasizes visual tokens consistently supporting high-confidence text generation.

The global attention map $\boldsymbol{a}_{\text{global}} \in \mathbb{R}^{3n}$ is obtained through element-wise summation of the three components: $\boldsymbol{a}_{\text{global}} = \boldsymbol{a}_{\text{self}} + \boldsymbol{a}_{\text{prompt}} + \boldsymbol{a}_{\text{text}}$. This comprehensive assessment is further averaged across the object dimension and normalized to get the final object-centric attention map $\boldsymbol{a} \in [0, 1]^{n}$, which serves as the training target for our GAP network.

\subsubsection{\textbf{Lightweight GAP Network Learning.}}

The GAP network learns to predict aggregated attention maps of the target 3D MLLM to facilitate visual token pruning. As shown in Figure \ref{fig:method}a, the network's input closely mirrors that of the target model, encompassing textual prompts and structured object-centric visual embeddings encoding the 3D scene. The network is designed to achieve precise and computationally efficient attention prediction:

\noindent{\textbf{Embedding Early Fusion.}} 
For each object $o_i$, we enhance spatial awareness by encoding its center coordinates $\boldsymbol{c}_i\in\mathbb{R}^3$ and bounding box dimensions $\boldsymbol{z}_i\in\mathbb{R}^3$ via linear projection to obtain the spatial location embedding as $\boldsymbol{l}_i = \mathbf{W}_l[\boldsymbol{c}_i;\boldsymbol{z}_i]\in \mathbb{R}^d$. An early fusion strategy is adopted to integrate visual embeddings from different sources. We first fuse 3D and 2D semantic embeddings with a multi-layer perceptron as $\mathbf{E}^s_i = \text{MLP}([\mathbf{E}^p_i; \mathbf{E}^v_i])$. Then, the object identifiers, fused semantic embeddings, and spatial location features are added together as the final fused object-centric embeddings: $\mathbf{F}_i = \mathbf{E}^o_i + \mathbf{E}^s_i + \boldsymbol{l}_i$. This compressed fusion scheme reduces the input sequence length while preserving multi-modal correlations.

\noindent{\textbf{Encoder-Decoder Architecture.}} The GAP network adopts a lightweight encoder-decoder architecture. We finetune a pre-trained BERT \cite{devlin2019bert} to encode the prompt into a sequence of text features. A spatial transformer decoder \cite{chen2022language} is then applied to learn cross-modal correlations between the text and object modality features. Each transformer layer consists of a spatial self-attention module, a cross-attention module and a feed-forward neural network. The spatial self-attention mechanism improves the understanding of spatial relations among objects referenced in the instruction. The final layer outputs attention logits normalized via softmax to produce predicted object-centric attention distribution $\boldsymbol{\hat{a}}$.

\noindent{\textbf{Multi-Objective Optimization.}} To optimize the GAP network, we employ a multi-objective loss function that simultaneously enforces attention fidelity and rank consistency, which are critical for effective downstream pruning:
\begin{gather}
\mathcal{L} = \underbrace{\text{KL}(\boldsymbol{a} \parallel \boldsymbol{\hat{a}})}_{\text{Attention Fidelity}} + \lambda \sum_{i,j} \underbrace{\mathbf{1}(a_i > a_j) \cdot \max(0, \hat{a}_j - \hat{a}_i + \delta)}_{\text{Rank Consistency}},
\label{equ:loss}
\end{gather}
where $\boldsymbol{a} \in [0, 1]^{n}$ represents the ground-truth attention map. The first term, based on Kullback-Leibler (KL) divergence, ensures precise attention distribution alignment while the second term explicitly enforces pairwise ranking consistency by penalizing violations in the relative ordering of attention scores. To prevent over-smoothing and improve discriminative capability, we apply temperature sharpening to $\boldsymbol{a}$ before KL computation.

\subsection{Sample-Adaptive Visual Token Pruning}
\label{sec:sap}

To improve the efficiency of the target 3D MLLM, we prune less important visual tokens based on the estimated token importance score $\boldsymbol{\hat{a}}$. Existing pruning strategies can be categorized into two types: 1) Uniform compression maintaining fixed retention ratios from an intermediate layer onward, such as FastV \cite{chen2025image} and SGL \cite{zhao2024stitch}; 2) Preserving different proportions of the most important visual tokens at each layer of the target model, exemplified by FitPrune \cite{ye2024fit}. These methods rigidly predefine pruning ratios through empirical validation or automated search, maintaining constant per-sample token budgets regardless of input characteristics during inference. However, this static approach fundamentally conflicts with \emph{the intrinsic variance in visual-linguistic complexity across input samples -- simpler instructions require fewer discriminative tokens while complex tasks demand richer representations.} To resolve this limitation, we propose Sample-Adaptive visual token Pruning (SAP), which dynamically adjusts layer-wise token budgets based on input complexity for improved efficiency.

Specifically, we hypothesize that the concentration degree of the predicted attention map $\boldsymbol{\hat{a}} \in [0, 1]^{n}$ correlates with sample difficulty: concentrated distributions indicate easier samples with clearer identification of key tokens, while uniform distributions suggest harder samples with obscure token significance. To operationalize this, we introduce layer-wise total attention thresholds $\theta_k$ to dynamically determine token retention counts, replacing fixed pruning ratios. Given an inference sample $x$ with predicted attention scores $\boldsymbol{\hat{a}}^{(x)}$ ranked in descending order via permutation $\boldsymbol{\pi}$, the number of retained visual tokens $r_k^{(x)}$ at layer $k$ (where each 3D MLLM triplet $\{\mathbf{V}^o_i, \mathbf{V}^p_i, \mathbf{V}^v_i\}$ is treated as a single token intuitively) can be calculated as:
\begin{gather}
r_k^{(x)} = \max \Big\{ j \in \{1,2,\ldots,n\} \,\Big|\, \sum_{i=1}^{j} \hat{a}_{\pi(i)}^{(x)} \leq \theta_k \Big\}.
\end{gather}
This formulation ensures ordered token retention based on predicted attention scores while enforcing cumulative threshold constraints. The dynamic computation of $\boldsymbol{\hat{a}}$ through the GAP network enables sample-specific retention ratios, achieving adaptive pruning at the instance level.

\begin{algorithm}[!t]
\caption{Pruning Strategy Searching}
\label{alg:pss}
\begin{algorithmic}[1]
\REQUIRE model $\mathcal{G}$, data batch \( \mathcal{X} \), predicted attention $\boldsymbol{\hat{a}}$, target FLOPs budget \( \delta \) and binary search parameters \( \epsilon,\alpha_{\text{min}},\alpha_{\text{max}} \)

\STATE Initialize \( \alpha_{\text{low}} = \alpha_{\text{min}} \), \( \alpha_{\text{high}} = \alpha_{\text{max}} \)
\STATE Initialize $\mathbf{P}^0 = [\theta_1^0, \theta_2^0, \cdots, \theta_L^0]$ manually or via FitPrune \cite{ye2024fit} as $\theta^0_k = \max_{x \in \mathcal{X}}\sum_{i=1}^{r_k^0} \hat{a}_{\pi(i)}^{(x)}$, where $r_k^0$ denotes the obtained static token retention count at layer $k$

\WHILE{\( \alpha_{\text{high}} - \alpha_{\text{low}} > \epsilon  \)}
    \STATE \( \alpha_{\text{mid}} \leftarrow (\alpha_{\text{low}} + \alpha_{\text{high}}) / 2 \)
    \STATE $C_{\text{mid}} \leftarrow \sum_{{x} \in \mathcal{X}} \Phi\big( \mathcal{G}, \mathbf{P}(\alpha_{\text{mid}}), \boldsymbol{\hat{a}}^{(x)}, {x} \big)$
    \STATE \textbf{ if } { \( C_{\text{mid}} \leq \delta \)}
        \textbf{ then } \( \alpha_{\text{low}} \leftarrow \alpha_{\text{mid}} \)
    \textbf{ else } \( \alpha_{\text{high}} \leftarrow \alpha_{\text{mid}} \)
\ENDWHILE
\STATE $\alpha^* \leftarrow \alpha_{\text{low}}$
\RETURN Pruning strategy $\mathbf{P}^*=[\alpha^*\theta^0_1, \alpha^*\theta^0_2, \cdots, \alpha^*\theta^0_L]$
\end{algorithmic}
\end{algorithm}
\setlength{\textfloatsep}{10pt}

Our SAP framework is fully parameterized by $\mathbf{P} = [\theta_1, \theta_2, \dots, \theta_L]$, representing the total attention thresholds for each layer. To determine an optimal pruning strategy under a given computational budget, we further introduce \textbf{Pruning Strategy Searching} (Algorithm \ref{alg:pss}). Specifically, we define the pruning strategy $\mathbf{P}$ as a scaled variant of a set of predefined baseline parameters $\mathbf{P}^0 = [\theta_1^0, \theta_2^0, \cdots, \theta_L^0]$, \textit{i.e.}, $\mathbf{P}(\alpha) = \alpha \mathbf{P}^0$, where $\mathbf{P}^0$ can be initialized manually or based on the FitPrune \cite{ye2024fit} algorithm. Given a small batch of validation data $\mathcal{X}$, the total computation overhead of the model $\mathcal{G}$ can be expressed as:
$f(\alpha) = \sum_{{x} \in \mathcal{X}} \Phi\big( \mathcal{G}, \mathbf{P}(\alpha), \boldsymbol{\hat{a}}^{(x)}, x \big).$
Since this cost function increases monotonically, we perform binary search to find the maximum $\alpha^*$ such that $f(\alpha) \leq \delta$, where $\delta$ denotes the predefined FLOPs computational budget. The optimal pruning strategy is then obtained as $\mathbf{P}^* = \alpha^* \mathbf{P}^0$. 

\begin{table*}[t]
\caption{Performance comparison with previous visual token compression methods. ``Pruning Ratio'' denotes the average ratio of pruned visual tokens. ``Score Ratio'' is obtained by calculating the average ratio of each score relative to Baseline.} 
\vspace{-2mm}
\label{tab:performance_comparison}
\resizebox{\linewidth}{!}{
\begin{tabular}{c|c|c|cc|cc|cc|cc|cc|c}
\toprule
 &
  \multirow{2}{*}{Method} &
  Pruning &
  \multicolumn{2}{c|}{ScanRefer} &
  \multicolumn{2}{c|}{Multi3DRefer} &
  \multicolumn{2}{c|}{Scan2Cap} &
  \multicolumn{2}{c|}{ScanQA} &
  \multicolumn{2}{c|}{SQA3D} &
  Score \\
 &         &  Ratio  & Acc@0.25 & Acc@0.5 & F1@0.25 & F1@0.5 & C@0.5 & B-4@0.5 & C & B-4 & EM   & EM-R & Ratio  \\ \midrule
\multirow{13}{*}{\rotatebox[origin=c]{90}{Chat-Scene \cite{huang2024chat}}}
 & Baseline   & 0\%   & 56.21 & 50.42 & 58.14 & 53.24 & 76.35 & 35.86 & 84.23 & 13.57 & 53.98 & 56.80 & 100.00\% \\ \cmidrule(lr){2-14}
 & \multirow{3}{*}{Random Pruning}     & 35\%  & 36.06 & 31.54 & 40.85 & 36.62 & 72.77 & 34.90 & 81.61 & 12.75 & 52.74 & 55.28 & 84.42\% \\ 
 &                             & 65\%  & 14.67 & 12.51 & 19.92 & 18.04 & 31.14 & 26.61 & 76.42 & 12.28 & 51.02 & 53.41 & 60.38\% \\ 
 &                             & 90\%  & 3.50 & 2.83 & 9.59 & 9.25 & 21.74 & 23.36 & 67.94 & 10.65 & 48.48 & 50.99 & 47.80\% \\  \cmidrule(lr){2-14}
 & \multirow{3}{*}{ToMe \cite{bolya2022token}}      & 35\%  & 50.43 & 45.39 & 51.95 & 47.70 & 73.43 & 35.76 & 83.46 & 13.39 & 52.64 & 55.12 & 94.69\% \\ 
 &                             & 65\%  & 26.86 & 24.48 & 30.99 & 28.98 & 45.10 & 29.11 & 83.28 & 13.03 & 52.22 & 54.81 & 73.24\% \\ 
 &                             & 90\%  & 3.72 & 2.96 & 9.63 & 9.28 & 21.45 & 23.05 & 67.03 & 10.51 & 47.83 & 50.31 & 47.31\% \\  \cmidrule(lr){2-14} 
 & \multirow{3}{*}{FastV \cite{chen2025image}}      & 35\%  & 55.40 & 49.86 & 57.39 & 52.69 & 75.94 & 35.89 & 84.49 & 12.90 & 54.22 & 56.77 & 99.04\% \\ 
 &                             & 65\%  & 29.13 & 26.55 & 33.61 & 31.44 & 47.42 & 30.60 & 84.96 & 13.40 & 53.27 & 55.91 & 76.55\% \\ 
 &                             & 90\%  & 3.92 & 3.23 & 9.20 & 8.91 & 18.65 & 21.88 & 65.28 & 10.31 & 48.16 & 50.76 & 46.34\% \\  \cmidrule(lr){2-14} 
 & \multirow{3}{*}{GAP (Ours)}       & 35\%  & 56.61 & 51.02 & 58.33 & 53.52 & 75.83 & 35.56 & 85.13 & 13.06 & 53.95 & 56.49 & 99.79\% \\ 
 &                             & 65\%  & 56.47 & 50.79 & 58.46 & 53.86 & 73.24 & 34.77 & 85.23 & 13.00 & 53.55 & 56.25 & 99.10\% \\ 
 &                             & 90\%  & 56.09 & 50.93 & 55.60 & 51.43 & 69.25 & 32.98 & 84.04 & 13.42 & 52.59 & 55.04 & 96.87\% \\  
\bottomrule 
\end{tabular}}
\vspace{-1mm}
\end{table*}

\section{Experiment}
\subsection{Experimental Setup}
\noindent \textbf{Datasets and Metrics.} We conduct experiments on five 3D vision-language datasets: ScanRefer \cite{chen2020scanrefer} for single-object visual grounding, Multi3DRefer \cite{zhang2023multi3drefer} for multi-object visual grounding, Scan2Cap \cite{chen2021scan2cap} for dense captioning, and both ScanQA \cite{azuma2022scanqa} and SQA3D \cite{ma2022sqa3d} for visual question answering. All experiments on these datasets follow the default settings and metrics. For ScanRefer \cite{chen2020scanrefer}, we evaluate thresholded accuracy with Acc@0.25 and Acc@0.5, where predictions are deemed positive if they exhibit higher Intersection over Union (IoU) with the ground truths than the thresholds. Multi3DRefer \cite{zhang2023multi3drefer} is assessed using the F1 score at IoU thresholds of 0.25 and 0.5. For Scan2Cap \cite{chen2021scan2cap}, we adopt CIDEr@0.5 and BLEU-4@0.5, integrating captioning metrics with the IoU scores. For ScanQA \cite{azuma2022scanqa}, we use CIDEr \cite{vedantam2015cider} and BLEU-4 \cite{papineni2002bleu}. SQA3D \cite{ma2022sqa3d} is measured using exact match accuracy (EM) and its refined version, EM-R \cite{huang2023embodied}.

\noindent \textbf{Implementation Details.} To validate Fast3D, we apply it to Chat-Scene \cite{huang2024chat} using the provided checkpoint. The base language model for this 3D MLLM is Vicuna-7B \cite{vicuna2023}. Under the default setting, Chat-Scene utilizes 300 object-centric visual tokens. Our experimental setup involves training a lightweight GAP network with a transformer-based encoder-decoder architecture, containing 159M total parameters. The GAP network undergoes unified training on combined datasets generated from all five benchmarks. We set the hidden dimension to 768 and use 12 attention heads for all the transformer layers. The language encoder is a three-layer transformer initialized from BERT \cite{devlin2019bert}, and the cross-modal decoder contains four layers. The hyper-parameter in the loss function Eq (\ref{equ:loss}) is set to $\lambda=0.02$. We train the model with a batch size of 64 and a learning rate of 0.0008 using cosine decay scheduling for 50 epochs. The AdamW algorithm \cite{loshchilov2017fixing} is employed in the optimization.

\subsection{Experimental Results}

\noindent \textbf{Effectiveness of Global Attention Maps.}
We first investigate the superiority of aggregating attention maps from all layers of the target 3D MLLM. We experiment using different sources to guide visual token pruning. All experiments are conducted based on Chat-Scene model, consisting of 32 layers. We prune 95\% visual tokens at the 2-nd layer without the sample-adaptive pruning mechanism, achieving average pruning ratio of 90\%. 

The results shown in Table \ref{table:attention_map_source} indicate that using the global attention maps aggregated from all layers outperforms using partial attention maps from specific layers, such as FastV \cite{chen2025image}. Moreover, the average performance consistently improves when the number of aggregated layers increases. This demonstrates that leveraging attention maps from multiple layers helps accurately retain essential visual tokens. Notably, using predicted attention maps from our GAP network achieves performance comparable to the global oracle of the target model, highlighting its effectiveness.

\begin{table}[t]
\centering
\caption{
Performance with attention maps from different sources. The visual token pruning ratio is set to 90\% for all experiments. Predicting attention maps via the GAP network achieves performance comparable to the global oracle of the target model.
}
\vspace{-1mm}
\resizebox{\columnwidth}{!}{%
\begin{tabular}{c|c|ccc|c}
\toprule
  \multicolumn{2}{c|}{attention map source}  & ScanRefer     & Scan2Cap   & SQA3D & score ratio \\ 
\midrule 
\multirow{6}{*}{\rotatebox[origin=c]{90}{Target Oracle}} & two layers (FastV \cite{chen2025image}) & 3.23 & 21.88 & 50.76 & 52.26\% \\
&10\% layers & 8.59 & 23.49 & 51.52 & 57.75\% \\
&30\% layers & 32.04 & 29.60 & 52.02 & 79.22\% \\
&50\% layers & 47.70 & 30.85 & 53.37 & 91.53\% \\
&70\% layers & 51.36 & 32.11 & 54.62 & 95.86\% \\
&all layers (global) & 52.72 & 33.45 & 55.10 & 98.28\%\\
\midrule
\multicolumn{2}{c|}{GAP Network (ours)} & 50.93 & 32.98 & 55.04 & 96.63\%\\
\bottomrule
\end{tabular}
}
\label{table:attention_map_source}
\end{table}

\noindent \textbf{Comparing GAP with Previous Methods.}
To validate the effectiveness of our GAP without the sample-adaptive pruning strategy, we present comparison results in Table \ref{tab:performance_comparison} with two representative visual token compression methods (\textit{i.e.}, ToMe \cite{bolya2022token} and FastV \cite{chen2025image}) and the random pruning, across different average visual token pruning ratios. For all methods, we apply object-centric pruning, treating each visual token triplet in the 3D MLLM as an integral unit, and adopt fixed pruning ratio from an intermediate layer onward. For our method, FastV \cite{chen2025image} and the random pruning, we prune 70\%, 80\% and 95\% visual tokens at the 16-th, 6-th, and 2-nd layer, achieving average pruning ratios of 35\%, 65\%, and 90\%, respectively. ToMe \cite{bolya2022token} performs token merging prior to the language model, with the merging ratio adjusted to achieve similar average pruning ratios. 

In dense captioning and visual question answering tasks, such as Scan2Cap \cite{chen2021scan2cap}, ScanQA \cite{azuma2022scanqa}, and SQA3D \cite{ma2022sqa3d}, when the token pruning ratio is relatively low (\textit{e.g.}, 35\%), all methods, even random pruning, exhibit performance comparable to the original model. This indicates significant visual token redundancy in object-centric 3D MLLMs, highlighting the importance of visual token pruning.

\begin{figure*}[!t]
    \centering
    \includegraphics[width=1\linewidth]{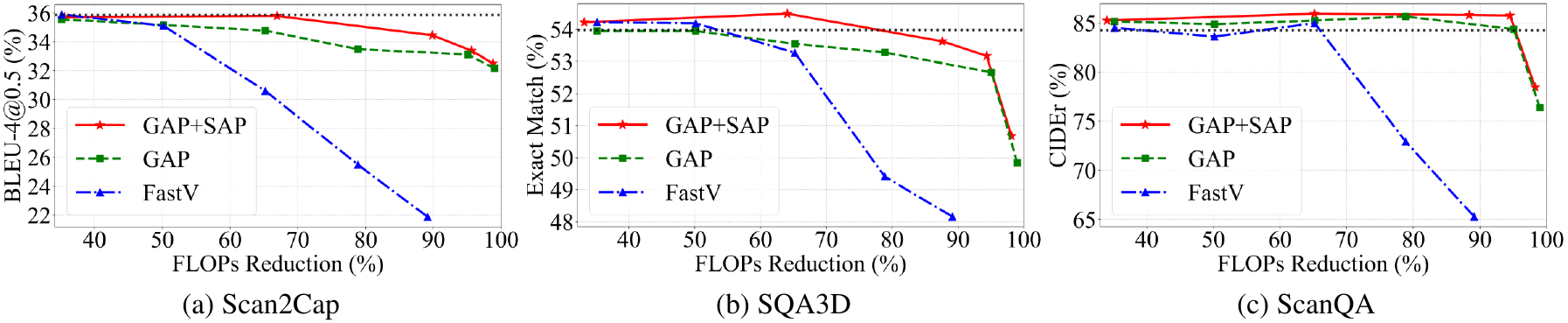}
    \vspace{-4mm}
    \caption{\textbf{Performance-efficiency trade-off curves of Fast3D (GAP + SAP).} The x-axis stands for the theoretical FLOPs reduction ratio under different configurations. The y-axis stands for performance under different settings. } 
    \label{fig:trade_off}
    \vspace{-2mm}
\end{figure*}

When the token pruning ratio is increased to 65\%, the performance of ToMe and FastV starts to drop, particularly in visual grounding tasks, including ScanRefer \cite{chen2020scanrefer} and Multi3DRefer \cite{zhang2023multi3drefer}, as well as the dense captioning task. These tasks require methods to accurately retain instruction- and answer-related visual tokens to understand 3D scene details. This performance decline shows that ToMe and FastV can not accurately retain essential tokens. In contrast, our method maintains competitive performance across all tasks.

With a visual token pruning ratio of 90\%, FastV and ToMe collapse across all tasks, exhibiting performance similar to random pruning, as critical visual tokens are lost due to inaccurate token importance estimation. In this challenging scenario, our method experiences only a marginal performance drop compared to other methods, achieving over 96.8\% of the original Chat-Scene's performance. This validates the superiority of our GAP, which successfully preserves the tokens most relevant to a correct answer, thanks to the global attention maps predicted from the GAP network.

\noindent \textbf{GAP with SAP Towards Improved Efficiency.}
We further validate the superiority of our Fast3D by incorporating both GAP and SAP mechanisms. The performance-efficiency trade-off curves for varying token pruning ratios across multiple benchmarks are illustrated in Figure \ref{fig:trade_off}. We report the overall performance alongside the corresponding theoretical FLOPs reduction ratios related to visual tokens. As detailed in Section \ref{sec:sap}, for each predefined computational budget, we search the optimal pruning strategy on validation samples and perform sample-adaptive pruning on test data based on the derived policy. Note that the additional computational overhead introduced by the GAP network accounts for merely 0.1\% of the target model's FLOPs related to visual tokens, rendering it negligible.

It can be observed that, at lower FLOPs reduction ratios, all methods maintain performance comparable to the original model. However, as the pruning ratio increases, our method GAP without SAP outperforms FastV by significant margins. Furthermore, the proposed SAP enables GAP to sustain performance without any degradation even at higher FLOPs reduction ratios. For instance, on Scan2Cap, the maximum reduction ratio at which performance remains competitive to the original model increases from 40\% to over 65\%, while on SQA3D, it rises from 50\% to approximately 80\%. These results demonstrate SAP's effectiveness in dynamically adjusting layer-wise token budgets based on input complexity, thereby achieving a better balance between overall performance and efficiency. 

In summary, our Fast3D provides superior advantages for visual token pruning in 3D MLLMs.

\begin{table}[!t]
\centering
\caption{
Performance of aggregating different attention maps as the GAP network's training target. 
}
\vspace{-2mm}
\resizebox{\columnwidth}{!}{%
\begin{tabular}{c|ccc|c}
\toprule
{training target}  & ScanRefer     & Scan2Cap   & SQA3D & score ratio \\ 
\midrule 
$\boldsymbol{a}_{\text{self}}$ & 29.34 & 27.37 & 51.76 & 75.21\% \\
$\boldsymbol{a}_{\text{prompt}}$ & 42.14 & 30.16 & 52.32 & 86.60\% \\
$\boldsymbol{a}_{\text{text}}$ & 46.11 & 31.25 & 52.74 & 90.48\% \\
$\boldsymbol{a}_{\text{self}}+\boldsymbol{a}_{\text{prompt}}$ & 44.68 & 31.85 & 54.37 & 91.05\% \\
$\boldsymbol{a}_{\text{prompt}}+\boldsymbol{a}_{\text{text}}$ & 49.47 & 32.75 & 54.49 & 95.12\% \\
$\boldsymbol{a}_{\text{self}}+\boldsymbol{a}_{\text{prompt}}+\boldsymbol{a}_{\text{text}}$ & \textbf{50.93} & \textbf{32.98} & \textbf{55.04} & \textbf{96.63}\%\\
\bottomrule
\end{tabular}
}
\label{table:different_attn_maps}
\vspace{-1mm}
\end{table}

\subsection{Ablation Studies}

\noindent \textbf{Key Attention Maps Aggregated as Target.} Our GAP network adopts aggregated multi-source attention scores as the training target, systematically integrating visual token self-attention $\boldsymbol{a}_{\text{self}}$, cross-modal attention with textual prompts $\boldsymbol{a}_{\text{prompt}}$ and generated text tokens $\boldsymbol{a}_{\text{text}}$. We ablate this design choice without SAP in Table \ref{table:different_attn_maps}. It can be observed that when using only a single attention map, cross-modal attention mechanisms, $\boldsymbol{a}_{\text{prompt}}$ and $\boldsymbol{a}_{\text{text}}$, outperform uni-modal self-attention $\boldsymbol{a}_{\text{self}}$, confirming the value of modality interaction. Furthermore, systematic integration of dual or triple attention maps yields consistent performance improvements, indicating complementary information across different attention maps. These findings validate that multi-source attention aggregation provides comprehensive visual token characterization through the combination of intra- and cross-modal relationships.

\begin{figure*}[!t]
    \centering
    \includegraphics[width=1.0\linewidth]{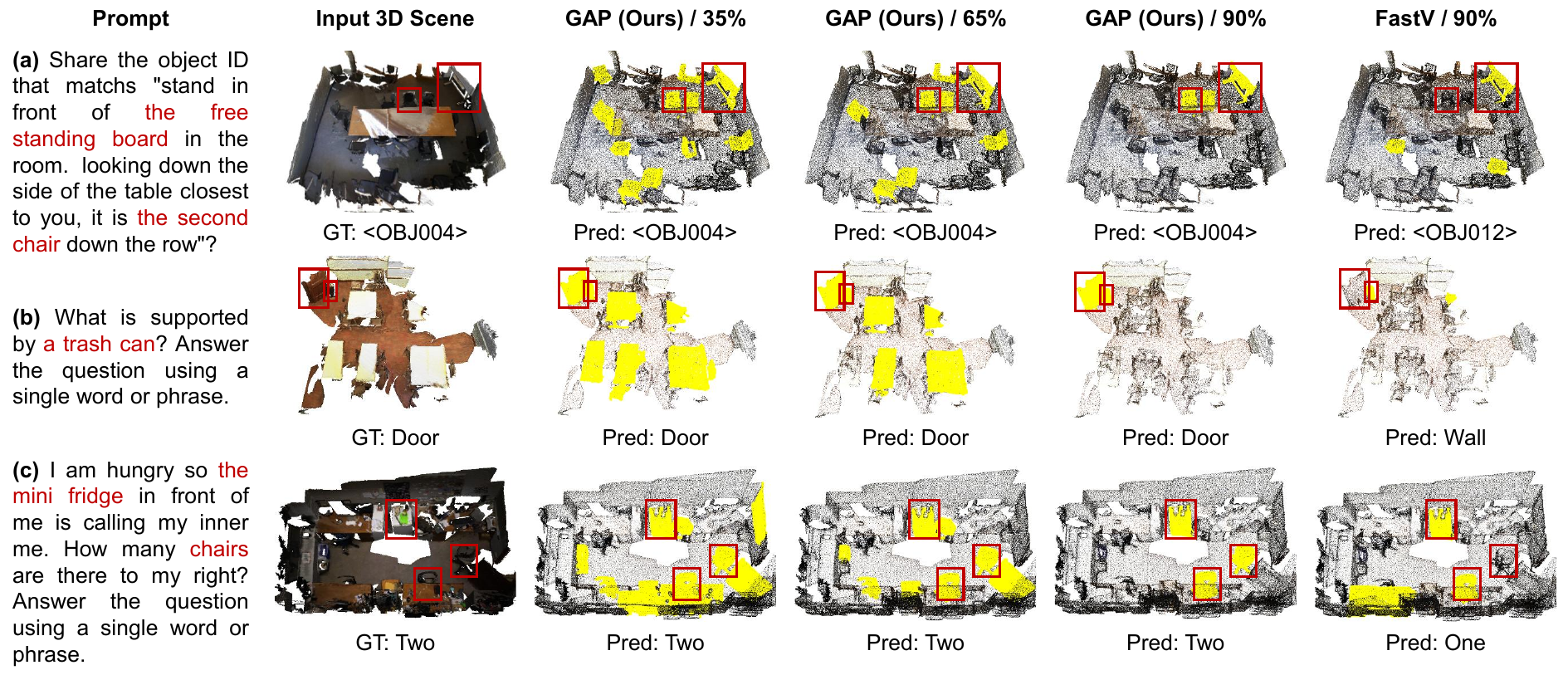}
    \vspace{-4mm}
    \caption{Visualization of GAP under different visual token pruning ratios. Object-centric visual tokens are pruned by 70\%, 80\%, and 95\% at the 16th, 6th, and 2nd layers of Chat-Scene, which comprises 32 layers. This results in average pruning ratios of 35\%, 65\%, and 90\%, respectively. Retained objects are marked in yellow. Red boxes highlight key objects referenced in prompts.} 
    \label{fig:visualization}
    \vspace{-1mm}
\end{figure*}

\noindent \textbf{Different Embeddings used in GAP Network.} We further investigate the effectiveness of different embeddings used in the embedding early fusion of our GAP network, including object identifiers, 3D and 2D semantic embeddings, and spatial location features. As shown in Table \ref{table:different_embeds}, object identifiers $\mathbf{E}^o$ emerge as indispensable components for the Scan2Cap dense captioning task, primarily due to their pivotal role in accurate object referring. 2D semantic features $\mathbf{E}^v$ derived from multi-view images yields better performance than using 3D features $\mathbf{E}^p$, suggesting that the pre-trained 2D encoder better capture discriminative object characteristics. Combining both 3D and 2D semantic embeddings achieve notable performance gain, validating the complementary nature of 3D and 2D information. Explicit integration of spatial location features $\boldsymbol{l}$ can further boost performance. These results demonstrate that fusing heterogeneous embeddings -- spanning identifier, semantic, and spatial domains -- provides comprehensive information for precise visual token importance estimation.

\begin{table}[!t]
\centering
\caption{
Performance of using different input embeddings in the GAP network under 90\% average pruning ratio. 
}
\vspace{-2mm}
\resizebox{\columnwidth}{!}{%
\begin{tabular}{c|ccc|c}
\toprule
{used embeddings}  & ScanRefer     & Scan2Cap   & SQA3D & score ratio \\ 
\midrule 
$\textbf{E}^p+\textbf{E}^v$ & 48.59 & 20.28 & 54.15 & 82.75\% \\
$\textbf{E}^o+\textbf{E}^p$ & 46.93 & 29.70 & 53.34 & 89.94\% \\
$\textbf{E}^o+\textbf{E}^v$ & 47.67 & 30.56 & 53.57 & 91.36\% \\
$\textbf{E}^o+\textbf{E}^p+\textbf{E}^v$ & 49.82 & 32.17 & 54.53 & 94.84\% \\
$\textbf{E}^o+\textbf{E}^p+\textbf{E}^v+\boldsymbol{l}$ & \textbf{50.93} & \textbf{32.98} & \textbf{55.04} & \textbf{96.63}\%\\
\bottomrule
\end{tabular}
}
\label{table:different_embeds}
\vspace{-1mm}
\end{table}

\noindent \textbf{Impact of Multi-Objective Optimization.} 
We further investigate the effects of using different loss functions to train our GAP network. While the KL divergence loss $\mathcal{L}_{\text{KL}}$ focus on absolute attention distribution matching, our proposed rank consistency loss $\mathcal{L}_{\text{rank}}$ emphasizes the relative ordering similarity of attention scores, making it more effective for downstream visual token pruning tasks where relative importance ranking determines preservation decisions. As shown in Table \ref{table:loss_functions}, although standalone KL divergence optimization yields competitive results, its combination with rank consistency loss produces significant improvements, validating the effectiveness of multi-objective optimization.

\subsection{Visualization Results}
To provide a deeper understanding of our GAP, we visualize the token pruning results in Figure \ref{fig:visualization}. Specifically, we present three representative examples from the ScanRefer, ScanQA, and SQA3D benchmarks. These results show that our GAP network consistently retains critical visual tokens essential for various 3D vision-language tasks across varying pruning ratios, demonstrating robust adaptability. Moreover, the network tends to retain core objects explicitly referenced in linguistic prompts (\textit{e.g.}, ``board'', ``chair'', ``trash can'', and ``fridge''), showing strong grounding capability to identify target objects. We also present the visual token pruning and answers from FastV \cite{chen2025image}. Our analysis reveals that FastV only preserves partial tokens relevant to the given instructions and struggles to retain them accurately in 3D scenes. Consequently, this limitation impairs the model’s ability to perceive key objects, leading to inaccurate predictions for complex spatial reasoning tasks.

\begin{table}[!t]
\centering
\caption{
Performance of using different loss functions to train the GAP network. 
}
\vspace{-2mm}
% \resizebox{\columnwidth}{!}{%
\begin{tabular}{c|ccc|c}
\toprule
{loss function}  & ScanRefer     & Scan2Cap   & SQA3D & score ratio \\ 
\midrule 
$\mathcal{L}_{\text{KL}}$ & 50.43 & 31.91 & 54.47 & 94.97\% \\
$\mathcal{L}_{\text{KL}}+\lambda\mathcal{L}_{\text{rank}}$ & \textbf{50.93} & \textbf{32.98} & \textbf{55.04} & \textbf{96.63}\%\\
\bottomrule
\end{tabular}
% }
\label{table:loss_functions}
\end{table}

\section{Conclusion}
In this paper, we explore visual token pruning for scene-level 3D MLLMs. Our findings reveal that token redundancy persists in object-centric representations and global attention remains an effective pruning signal in 3D domains. Based on these insights, we propose Fast3D, a plug-and-play visual token pruning framework tailored for 3D MLLMs with two key components. First, our Global Attention Prediction (GAP) mechanism employs a lightweight neural network to approximate the aggregated attention map from all layers of the target model, enabling efficient token importance estimation for precise pruning guidance. Second, the Sample-Adaptive visual token Pruning (SAP) strategy dynamically adjusts layer-wise pruning ratios based on input complexity, yielding improved overall accuracy-efficiency tradeoffs. Both components operate without modifying the target model's parameters. Experiments on five benchmarks validate the effectiveness of Fast3D, particularly under high visual token pruning ratios.

% \begin{acks}
% \end{acks}

\bibliographystyle{ACM-Reference-Format}
\balance
\bibliography{main}

% \appendix

\end{document}